\documentclass[9pt, conference]{IEEEtran}

\IEEEoverridecommandlockouts

\usepackage{cite}
\usepackage{amsmath,amssymb,amsfonts}
\usepackage{algorithm}
\usepackage{multirow}
\usepackage{algpseudocode}
\usepackage{graphicx}
\usepackage{textcomp}
\usepackage{xcolor}
\def\BibTeX{{\rm B\kern-.05em{\sc i\kern-.025em b}\kern-.08em
    T\kern-.1667em\lower.7ex\hbox{E}\kern-.125emX}}

\begin{document}
\setlength{\abovedisplayskip}{2.5pt}
\setlength{\belowdisplayskip}{2.5pt}

\title{FedTLU: Federated Learning with Targeted Layer Updates
\thanks{The authors are partially supported by the National Science Foundation under grants CNS-2409138 and CNS-2106891 and by the Department of Energy under grant DESC0025652.}

}

\author{
\IEEEauthorblockN{Jong-Ik Park}
\IEEEauthorblockA{\textit{Department of Electrical and Computer Engineering} \\
\textit{Carnegie Mellon University}\\
Pittsburgh, USA \\
jongikp@andrew.cmu.edu}
\and
\IEEEauthorblockN{Carlee Joe-Wong}
\IEEEauthorblockA{\textit{Department of Electrical and Computer Engineering} \\
\textit{Carnegie Mellon University}\\
Pittsburgh, USA \\
cjoewong@andrew.cmu.edu}
}

\maketitle

\begin{abstract}
Federated learning (FL) addresses privacy concerns in training language models by enabling multiple clients to contribute to the training, without sending their data to others. However, non-IID (identically and independently distributed) data across clients often limits FL's performance. This issue is especially challenging during model fine-tuning, as noise due to variations in clients' data distributions can harm model convergence near stationary points. This paper proposes a targeted layer update strategy for fine-tuning in FL. Instead of randomly updating layers of the language model, as often done in practice, we use a scoring mechanism to identify and update the most critical layers, avoiding excessively noisy or even poisoned updates by freezing the parameters in other layers. We show in extensive experiments that our method improves convergence and performance in non-IID settings, offering a more efficient approach to fine-tuning federated language models.
\end{abstract}


\section{Introduction}
Data-driven language models have recently driven major application advancements in natural language processing like conversational agents and machine translation~\cite{vaswani2017attention, devlin2018bert, radford2019language, yao2024survey}. However, the widespread use of language data to train these models raises privacy concerns, as language data from individuals is personal and prone to exploitation for malicious purposes\cite{wu2020fedmed, hilmkil2021scaling, yao2024survey}.

Federated Learning (FL) addresses these concerns by training a \textit{global model} without centralizing user data~\cite{mcmahan2017communication, li2020federated, li2020preserving}. 
However, one of the core challenges in FL is developing a global model that performs well across clients, whose data can vary significantly in size and distribution~\cite{wang2021federated, qu2022generalized}.
This non-IID nature of client data often causes federated models to underperform compared to centralized models~\cite{abdulrahman2020survey, rahman2020internet, wang2021federated}. In language models, this heterogeneity can manifest as clients' exhibiting distinct linguistic patterns, content, and topics, complicating model convergence and generalization~\cite{ganiz2010higher, cao2022beyond, wu2024fedbiot}. Toward the end of the training, as the model nears a stationary point, stochastic local updates—amplified by data heterogeneity—create variability that makes it difficult to maintain global model consistency~\cite{cutkosky2019momentum, dinh2020federated}. This variability slows convergence and prevents the global model from reaching optimal performance~\cite{cutkosky2019momentum, dinh2020federated, liu2022few, he2023sensitivity}.

While local fine-tuning helps models adapt to client-specific data~\cite{li2021ditto, tan2022towards}, fine-tuning only local models can lead to divergence, as each client’s model becomes too specialized to its local data and tends to overfit, reducing the global model’s overall effectiveness across all clients~\cite{bietti2022personalization}. 
Moreover, individual clients often lack enough data for effective independent fine-tuning~\cite{mendieta2022local, ye2023heterogeneous}. A well-optimized global model, updated across clients, provides a stronger starting point~\cite{park2024fedbaf}. Once aligned, local fine-tuning can enhance specific client tasks without diverging from the global objective. Therefore, FL requires a well-generalized global model to meet the collective needs of all participants. Thus, our primary focus in this work is on fine-tuning the global model to ensure consistency across clients, with local tuning as an optional step for personalization.

In FL, a central server periodically aggregates locally updated client models. However, uniformly updating all model layers during aggregation, especially in the final stages, can amplify noise from certain clients, degrading performance and hindering convergence~\cite{liu2022threats, fang2022robust}. Toward the end of training, as the model nears a stationary point and gradients shrink, noisy or non-representative updates from certain clients can disproportionately affect the global model~\cite{li2020federated}. 
We can mitigate the impact of such noisy updates by selectively updating only the most important parameter sets, such as layers, which are likely to significantly reduce global loss. However, the key challenge in this approach is \textit{determining which layers to update}. Clients do not know the updates made by other clients and have limited information about the criticality of each layer. Therefore, server-side coordination is needed to aggregate client information and determine the most critical layers to update, though the server has only limited knowledge of the heterogeneous data distributions across clients. 

In this context, we propose Federated Learning with Targeted Layer Updates (FedTLU). FedTLU selectively identifies and updates only the most critical layers to reduce global loss on the server side. By focusing on these key layers, FedTLU enables more efficient training, faster convergence, and improved global model performance across diverse clients.
In this work, we make three primary \textbf{contributions}: \\
$\bullet$ We propose a novel \textit{server-side scoring mechanism} that identifies critical layers to update without relying on client-side knowledge. \\
$\bullet$ FedTLU is designed to be \textit{robust against noisy updates}, even in environments with non-representative or noisy client data. \\
$\bullet$ \textit{FedTLU consistently outperforms random and last-layer updates}, achieving better test performance by up to 7.86\% globally and up to 8.27\% locally, particularly during the noisy fine-tuning phase. 

We review related work in Section~\ref{sec:related}, present the FedTLU algorithm in Section~\ref{sec:method}, and offer theoretical support for FedTLU's design in Section~\ref{sec: analysis}. Section~\ref{sec:experiments} presents experimental results that validate our approach, and we summarize our findings in Section~\ref{sec:conclusion}.
\section{Related Work}\label{sec:related}
In both centralized learning (CL) and FL, fine-tuning is widely recognized for improving model performance, particularly during the final stages of training. In CL, where all data is stored on a central server, fine-tuning often involves updating only a small subset of model parameters, such as layers near the output nodes~\cite{kenton2019bert}, or even selecting layers randomly for updating~\cite{li2016improved, swiderski2022random, goutam2020layerout}. This selective approach improves convergence and reduces computational costs~\cite{raghu2019transfusion, houlsby2019parameter}. Refining a subset of layers allows the model to adjust task-specific representations while leveraging previously learned features~\cite{li2020preserving, yu2021differentially}.
These layers also can be selected based on their importance, determined through explainable neural network techniques like sensitivity or criticality scoring that assess how vital certain parameters are for generalization~\cite{novak2018sensitivity, kumar2022corrnet}. Such methods prioritize updates where they will have the most significant impact.

Similarly, fine-tuning a subset of layers in FL can benefit convergence. FL can theoretically mirror CL under certain conditions, particularly if there are enough communication rounds between clients and the central server~\cite{li2020federated, karimireddy2020scaffold}. Thus, some fine-tuning techniques from CL can be adapted for FL scenarios~\cite{tan2022towards, kairouz2021advances}. However, CL techniques for scoring layers are difficult to extend to FL: in FL, clients lack information about others' updates, and the server can only aggregate what is provided.  
As a result, simpler strategies like randomly selecting layers or updating the last few layers are often used~\cite{tian2022fedbert}. However, fine-tuning only the last few layers is insufficient in non-IID settings, where data heterogeneity leads to highly varied learned representations~\cite{zhu2021federated, smith2017federated, sattler2020clustered}. This variability can hinder the global model's generalization ability, as the last layer represents drastically different outputs depending on each client’s data. We therefore focus on updating the main body of the model. 

We aim to update the most significant layers contributing to reducing global loss, where each layer’s update is guided by the local gradient magnitudes and directions during local training, resulting in varied contributions from different layers~\cite{morcos2018importance, li2021ditto, liu2022threats}. Rather than random selection, we enhance model stability by selecting impactful layers. By targeting key layers that effectively reduce global loss, we can improve convergence, especially in language tasks, addressing the limitations of random or fixed-layer updates~\cite{li2019convergence}.

\section{Methodology}\label{sec:method}
This section introduces FedTLU, a method for selecting layers to update during global aggregation in FL, particularly for training language models with non-IID data. The goal is to efficiently update the most critical layers or blocks to improve model convergence and generalization, especially in the final training stages. 

We assume the usual FL training framework, in which training proceeds iteratively and in each round $t$, clients first compute local model updates based on their local data. These updated model parameters are then sent to the server, which aggregates them to form the new global model with parameters \(W_{\text{aggre},i}^{(t+1)}\) in each layer $i$.

\textbf{Layer scoring:} To quantify the significance of each layer, we define a score for the weights \(W_i\) in each layer $i$ as follows:
\[
\text{Score}(W_i) = \frac{\|\Delta W_i\|}{\sqrt{n_i} \cdot \text{std}(\Delta W_i)},
\]
where \(\|\Delta W_i\|\) is the $L2$ norm of the difference between the parameters before round $t$'s update \((W_i^{(t)})\) and the aggregated parameters after the update \(W_{\text{aggre},i}^{(t+1)}\), \(n_i\) is the number of parameters in layer \(W_i\), and \(\text{std}(\Delta W_i)\) is the standard deviation of the differences in parameters in layer $i$. This score is large in two cases:

\paragraph{Large Numerator (\(\|\Delta W_i\|\))} A large numerator indicates substantial changes between the current and aggregated parameters, suggesting that the layer is making a significant contribution to reducing the global loss~\cite{li2023zico}.

\paragraph{Small Denominator (\(\sqrt{n_i} \cdot \text{std}(\Delta W_i)\))} A small standard deviation \(\text{std}(\Delta W_i)\) reflects consistent parameter changes, indicating an effective gradient update. The factor \(\sqrt{n_i}\) normalizes by layer size, allowing for fair comparison across layers~\cite{morcos2018importance, li2023zico}.

\textbf{Training algorithm:} Modern deep neural networks, particularly language models, often consist of repeated blocks of layers with the same sequence of channel sizes~\cite{vaswani2017attention, radford2019language}. While individual layers may differ in parameters, repeated blocks typically have the same total parameter count. To leverage this structure, we group layers into blocks for comparison and updating, so that we can fairly compare blocks with similar roles. By using aggregated scores for these blocks, we account for gradient magnitude and consistency.

\begin{algorithm}[t]
\caption{This algorithm presents the process of FedTLU. Number of clients $K$, learning rate $\eta$, number of global rounds $T$ after pre-training \(T_0\), number of selected blocks $S$, local epochs $E$, participation rate $\mathcal{C}$}
\label{alg:flex-block-fed}
\begin{algorithmic}[1]
\State \textbf{Input:} Pre-trained global model $W^{(T_0)}$, participation rate $\mathcal{C}$
\State \textbf{Output:} Final global model $W^{(T)}$
\For{each round $t = T_0, \ldots, T$}
    \State Server randomly selects $\mathcal{C} K$ clients from $K$ clients
    \State Server sends $W^{(t)}$ to the selected clients
    \For{each client $k \in \{\text{Selected Clients}\}$ \textbf{in parallel}}
        \State $W^{(t+1)}_k \gets$ \Call{LocalUpdate}{$W^{(t)}$, $E$, $\eta$}
        \State Client $k$ sends $W^{(t+1)}_k$ to the server
    \EndFor   
    \State $W^{(t+1)}_{\text{aggre}} \gets$ \Call{Aggregation}{$\{W^{(t+1)}_k\}_{k \in \{\text{Selected Clients}\}}$}
    \State $\text{LayerScores} \gets$ \Call{ComputeLayerScore}{$W^{(t)}$, $W_{\text{aggre}}^{(t+1)}$}
    \State $\text{BlockScores} \gets$ \Call{ComputeBlockScore}{$\text{LayerScores}$}
    
    \State Group blocks by the sequence of parameters
    \For{each group $G_i$}
        \State $\text{SelBlocks} \gets$ \Call{SelectBlocks}{$\text{BlockScores}$ in $G_i$, $S$}    
    \EndFor
    \State Global model $W^{(t+1)}$ is updated with SelBlocks from each group in $W^{(t+1)}_{\text{aggre}}$
\EndFor
\vspace{0.1cm}
\Function{SelectBlocks}{$\{\text{Score}(B_i)\}$, $S$}
    \State \text{SortedB} $\gets$ Sort blocks by \text{Score}$(B_i)$
    \State \text{SelectedB} $\gets$ Select the top $S$ blocks from \text{SortedB}
    \State \Return \text{SelectedB}
\EndFunction
\end{algorithmic}
\end{algorithm}

During each round of aggregation, the process begins by grouping blocks with the same number of parameters into sets, allowing for structured comparison. For each block \(B_i\), an aggregated score is computed by summing the scores of all layers within the block:
\[
\text{Score}(B_i) = \sum_{j \in B_i} \text{Score}(W_j).
\]
Algorithm~\ref{alg:flex-block-fed} outlines the FedTLU process. In each global round (\textit{line 2}), the server randomly selects a subset of clients based on the participation rate $\mathcal{C}$ (\textit{line 3}). Selected clients receive the global model $W^{(t)}$ and perform local updates (\textit{line 6}), then return their updated models to the server (\textit{line 7}). The server aggregates these models (\textit{line 9}) and computes a score for each layer based on the difference between current and aggregated parameters (\textit{line 10}). Aggregation methods can vary, including FedAvg~\cite{mcmahan2017communication} or FedProx~\cite{li2020federated}.

Block scores are computed by summing the scores of all layers within each block (\textit{line 11}), and blocks are grouped by their parameter sequence for fair comparison (\textit{line 13}). The top $S$ blocks from each group are selected based on their scores (\textit{line 14}), and the global model is updated with these blocks (\textit{line 16}). This process repeats for $T$ global rounds, focusing updates on the most important layers or blocks to improve convergence across heterogeneous clients.

\section{Theoretical Analysis} \label{sec: analysis}
In this section, we theoretically analyze why updating only a few layers of the global model during aggregations can benefit FL.
\subsection{Assumptions}
We make the following assumptions in our analysis:
\paragraph{Gradient Smoothness} 
The gradients of the global loss function are Lipschitz continuous, implying that the global loss function is smooth. Specifically, for any aggregated global model \( W \), the loss function \( \mathcal{L}(W) \) is \( L \)-smooth, meaning:
\[
\mathcal{L}(W') \leq \mathcal{L}(W) + \langle \nabla \mathcal{L}(W), W' - W \rangle + \frac{L}{2} \|W' - W\|^2,
\]
where \( L \) is the Lipschitz constant, and \( \nabla \mathcal{L}(W) \) is the gradient of the global loss function with respect to the model \( W \).

\paragraph{Effective Gradient in Subset \(W_\mathcal{S}^{(t)}\)} 
We assume that the subset of layers \(W_\mathcal{S}^{(t)}\) contains layers where the gradient most effectively reduces the loss. Specifically, we assume that in each round $t$:
\[
\|\nabla_{W_\mathcal{S}^{(t)}} \mathcal{L}(W^{(t)})\|^2 \geq \|\nabla \mathcal{L}(W^{(t)})\|^2 - \delta,
\]
where \( \delta \) is a positive constant. While smaller \(\delta\) values indicate better alignment with the full model’s gradient, a larger \(\delta\) can be advantageous in noisy or complex scenarios, as perfect alignment may lead to updating every source of noise in the model.

\paragraph{Client Participation and Data Distribution}
For simplicity, we assume full participation from all clients in every round of communication. To remove stochasticity from the analysis, we also assume that each client's local update is computed with one step of gradient descent with the full gradient of its local loss over all of its data in every round. Additionally, we assume that each client’s local data distribution remains constant throughout the training process.

\subsection{Analysis of Gradient Descent Updates}
We now compare the impact of updating all layers versus updating only a subset of layers during the global aggregation phase.

\paragraph{Full Model Update}
When the entire model \( W \) is updated during the global aggregation phase, the update rule for each layer \( W_i \) at iteration \( t \) is 
\(
W_i^{(t+1)} = W_i^{(t)} - \eta \nabla_{W_i} \mathcal{L}(W^{(t)}),
\)
where \( \nabla_{W_i} \mathcal{L}(W^{(t)}) = \sum_{k=1}^{K} p_k \nabla_{W_i} \mathcal{L}_k(W^{(t)}) \). The overall change in the global loss function due to this update is governed by the smoothness condition, which can be expressed as:
\[
\mathcal{L}(W^{(t+1)}) - \mathcal{L}(W^{(t)}) \leq -\eta \|\nabla \mathcal{L}(W^{(t)})\|^2 + \frac{\eta^2 L}{2} \|\nabla \mathcal{L}(W^{(t)})\|^2.
\]
where \( \eta \) is the learning rate, \( \|\nabla \mathcal{L}(W^{(t)})\|^2 \) represents the squared $L2$ norm of the global gradient \( \nabla \mathcal{L}(W)^{(t)} \), and \( L \) is the Lipschitz constant of the global loss function. 

\paragraph{Subset of Layers Update} 
Now, we consider updating only a subset of layers, \( \mathcal{S}\). We decompose the model \( W^{(t)} \) into two parts: \\
$\bullet$ \( W_\mathcal{S}^{(t)} \): The subset of layers that are being updated. \\
$\bullet$ \( W_{\mathcal{S}^c}^{(t)} \): The layers that are not being updated. \\
Thus, without loss of generality the full model can be expressed as \( W^{(t)} = (W^{(t)}_\mathcal{S}, W^{(t)}_{\mathcal{S}^c}) \), where \( \mathcal{S}^c \) denotes the complement of \( \mathcal{S} \). Since the update only affects \( W_\mathcal{S} \), the difference \( W^{(t+1)} - W^{(t)} \) simplifies to : \( (W_\mathcal{S}^{(t+1)} - W_\mathcal{S}^{(t)}, 0)\).
Substituting into the smoothness condition:
\begin{equation*}
    \begin{aligned}
        \mathcal{L}(W^{(t+1)}) \leq& \mathcal{L}(W^{(t)}) + \langle \nabla \mathcal{L}(W^{(t)}), (W_\mathcal{S}^{(t)} - W_\mathcal{S}^{(t+1)}, 0) \rangle \\
        &+ \frac{L}{2} \|(W_\mathcal{S}^{(t+1)} - W_\mathcal{S}^{(t)}, 0)\|^2.
    \end{aligned}
\end{equation*}

Now, let’s focus on the gradient \( \nabla \mathcal{L}(W^{(t)}) \). This gradient is with respect to the entire model \( W^{(t)} \), which means it includes components for both \( W_\mathcal{S}^{(t)} \) and \( W_{\mathcal{S}^c}^{(t)} \), leading to \( \langle \nabla \mathcal{L}(W^{(t)}), (W_\mathcal{S}^{(t+1)} - W_\mathcal{S}^{(t)}, 0) \rangle = \langle \nabla_{W_\mathcal{S}^{(t)}} \mathcal{L}(W^{(t)}), W_\mathcal{S}^{(t+1)} - W_\mathcal{S}^{(t)} \rangle \).

Here, \( \nabla_{W_\mathcal{S}^{(t)}} \mathcal{L}(W^{(t)}) = \frac{\partial \mathcal{L}(W^{(t)})}{\partial W_\mathcal{S}^{(t)}} \) is the gradient of the loss function with respect to the subset \( W_\mathcal{S}^{(t)} \). Substituting this into the smoothness condition gives:
\begin{equation*}
    \begin{aligned}
    \mathcal{L}(W^{(t+1)}) - \mathcal{L}(W^{(t)}) \leq & \langle \nabla_{W_\mathcal{S}^{(t)}} \mathcal{L}(W^{(t)}), W_\mathcal{S}^{(t+1)} - W_\mathcal{S}^{(t)} \rangle\\ & + \frac{L}{2} \|W_\mathcal{S}^{(t+1)} - W_\mathcal{S}^{(t)}\|^2.
    \end{aligned}
\end{equation*}

\subsection{Theoretical Comparison: Full Model vs. Subset Updates}
Let us assume that the gradient over the subset \( W_\mathcal{S}^{(t)} \) is not aligned with the full gradient, i.e., large $\delta$.
The loss reduction is defined as:
\[
\Delta = \mathcal{L}(W^{(t)})-\mathcal{L}(W^{(t+1)}).
\]
For the subset update, the loss reduction can be expressed as:
\[
\Delta_{\mathcal{S}} \geq \eta \left(\|\nabla \mathcal{L}(W^{(t)})\|^2 - \delta \right) \left(1 - \frac{\eta L}{2} \right).
\]
For the full model update, the loss reduction is:
\[
\Delta_\text{all} \geq \eta \|\nabla \mathcal{L}(W^{(t)})\|^2 \left(1 - \frac{\eta L}{2} \right).
\]

Given the loss reduction inequalities for subset and full model updates, consider the scenario where the subset update yields a greater loss reduction than the full model update. This implies that the subset update provides a tighter boundary. The inequality describing the difference between the loss reduction boundaries for the subset update and the full model update is given by:
\[
-\eta \delta \left(1 - \frac{\eta L}{2} \right)  \geq 0.
\]
For this inequality to hold, we require \(1 - \frac{\eta L}{2} \leq 0\), which implies \(\eta L \geq 2\). This condition indicates that when the Lipschitz constant \(L\) is large, even a relatively small learning rate \(\eta\) causes the loss function to behave less smoothly, meaning the updates may become noisy. In such cases, deviating from the full model with a large \(\delta\) update can be beneficial, allowing for more efficient loss reduction by focusing on key layers.
\begin{table*}[ht] \label{tab: from_scratch}
\centering
\caption{Averaged minimum global and local test perplexities 
when the global model was trained from scratch.
}
\setlength{\tabcolsep}{0.5pt}
\begin{tabular}{|c|c|ccc|ccc|ccc|ccc|ccc|ccc|}
\hline
\multirow{2}{*}{Model} & Aggre  & \multicolumn{9}{c|}{FedAvg}  & \multicolumn{9}{c|}{FedProx} \\ 
\cline{2-20} 
\multirow{2}{*}{(Dataset)} & Portion  & \multicolumn{3}{c|}{0.75} & \multicolumn{3}{c|}{0.50} & \multicolumn{3}{c|}{0.25} & \multicolumn{3}{c|}{0.75} & \multicolumn{3}{c|}{0.50} & \multicolumn{3}{c|}{0.25} \\ 
\cline{2-20} 
& Update & \textbf{Ours} & Random & Last & \textbf{Ours} & Random & Last & \textbf{Ours} & Random & Last & \textbf{Ours} & Random & Last & \textbf{Ours} & Random & Last & \textbf{Ours} & Random & Last \\
\hline
Transformer & Global & \textbf{2.107} & 2.108 & 4.652 & \textbf{2.399} & 2.406 & 4.652 & \textbf{2.874} & 2.879 & 4.652 & \textbf{2.114} & 2.114 & 4.665 & \textbf{2.407} & 2.413 & 4.665 & \textbf{2.883} & 2.888 & 4.665 \\
(Penn TB) & Local & \textbf{2.109} & 2.110 & 4.559 & \textbf{2.399} & 2.406 & 4.559 & \textbf{2.870} & 2.876 & 4.559 & \textbf{2.116} & 2.116 & 4.572 & \textbf{2.407} & 2.414 & 4.572 & \textbf{2.880} & 2.885 & 4.572 \\
 \hline
GPT-2 & Global & \textbf{10.091} & 10.124 & 14.832 & \textbf{10.231} & 10.295 & 14.832 & \textbf{10.464} & 10.583 & 14.832 & \textbf{10.093} & 10.125 & 14.838 & \textbf{10.232} & 10.297 & 14.838 & \textbf{10.466} & 10.585 & 14.838 \\
(UDPOS) & Local & \textbf{10.245} & 10.287 & 12.016 & \textbf{10.361} & 10.415 & 12.016 & \textbf{10.560} & 10.695 & 12.016 & \textbf{10.247} & 10.288 & 12.018 & \textbf{10.362} & 10.416 & 12.018 & \textbf{10.563} & 10.696 & 12.018 \\
\hline
\end{tabular}
\end{table*}

\begin{table}[ht] \label{tab: dynamic}
\centering
\caption{Averaged minimum global and local test perplexities
when the global model was fine-tuned after 400 rounds.}
\setlength{\tabcolsep}{1.3pt}
\begin{tabular}{|c|c|cccc|cccc|}
\hline
Model & Aggre & \multicolumn{4}{c|}{FedAvg} & \multicolumn{4}{c|}{FedProx} \\ 
\cline{2-10} 
(Dataset) & Update & Full & \textbf{Ours} & Random & Last & Full & \textbf{Ours} & Random & Last \\
\hline
Transformer & Global & 1.774 & \textbf{1.730} & 1.730 & 1.785 & 1.776 & \textbf{1.731} & 1.732 & 1.787 \\
(Penn TB) & Local & 1.776 & \textbf{1.731} & 1.732 & 1.787 & 1.778 & \textbf{1.733} & 1.734 & 1.789 \\
 \hline
GPT-2 & Global & 9.617 & \textbf{9.292} & 9.382 & 9.645 & 9.618 & \textbf{9.293} & 9.383 & 9.646 \\
(UDPOS) & Local & 9.702 & \textbf{9.406} & 9.493 & 9.722 & 9.702 & \textbf{9.407} & 9.494 & 9.722 \\
\hline
\end{tabular}
\end{table}
\begin{table}[ht] \label{tab: dynamic_attack}
\centering
\caption{Averaged minimum global and local test perplexities 
when the models were fine-tuned with noisy or malicious clients}
\setlength{\tabcolsep}{1.3pt}
\begin{tabular}{|c|c|ccc|ccc|}
\hline
Model & Aggre & \multicolumn{3}{c|}{FedAvg} & \multicolumn{3}{c|}{FedProx} \\ 
\cline{2-8} 
(Dataset) & Update & Full & \textbf{Ours} & Random & Full & \textbf{Ours} & Random \\
\hline
Transformer & Global & 1.815 & \textbf{1.768} & 1.764 & 1.817 & \textbf{1.770} & 1.766 \\
(Penn TB) & Local & 1.804 & \textbf{1.765} & 1.763 & 1.806 & \textbf{1.767} & 1.764 \\
 \hline
GPT-2 & Global & 9.597 & \textbf{9.318} & 10.050 & 9.763 & \textbf{9.559} & 10.050 \\
(UDPOS) & Local & 9.696 & \textbf{9.444} & 10.225 & 9.869 & \textbf{9.687} & 10.225 \\
\hline
\end{tabular}
\end{table}

\begin{figure*}[ht] \label{fig: dynamic graph}
    \centering
    \includegraphics[width=\linewidth]{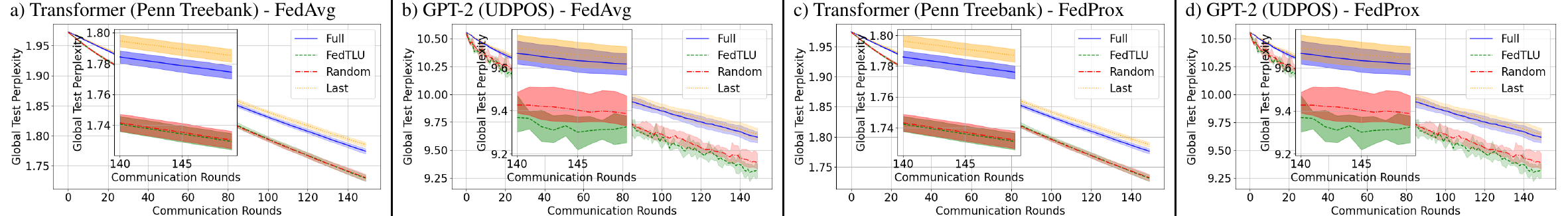}
    \vspace{-0.5 cm}
    \caption{Perplexity curves for fine-tuning with Full, FedTLU, Random, and Last-layer updates on Transformer (Penn TB) and GPT-2 (UDPOS) models.}
\end{figure*}

\begin{figure*}[ht] \label{fig: dynamic attack graph}
    \centering
    \includegraphics[width=\linewidth]{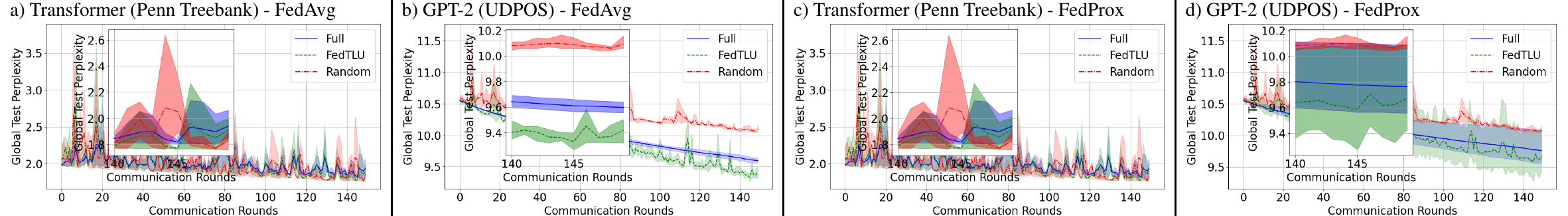}
    \vspace{-0.5 cm}
    \caption{Perplexity curves for fine-tuning with Full, FedTLU, and Random updates in noisy or malicious client scenarios.}
\end{figure*}

\section{Experimental Evaluation} \label{sec:experiments}
\subsection{Experimental Setup and Evaluation Metrics}
We trained two language models for next-word prediction: a standard Transformer~\cite{vaswani2017attention} on UDPOS data and GPT-2~\cite{radford2019language} on Penn Treebank (Penn TB) for 150 communication rounds. The sequence length was set to 128. Both models were optimized with Stochastic Gradient Descent (SGD) at a learning rate of 0.0001 and batch size of 16, with each client performing 5 local epochs across 100 clients. We used FedAvg~\cite{mcmahan2017communication} and FedProx~\cite{li2020federated} (proximal term $\mu=0.1$) for aggregation. Experiments were conducted on a single NVIDIA GeForce RTX 2080 Ti.

The network architectures consist of the repeated (main body) and non-repeated parts (input and output layers). Four update strategies were evaluated: \textbf{Full}, \textbf{FedTLU}, \textbf{Random}, and \textbf{Last} updates. In the Full update, all layers were updated in each round. For Random and FedTLU, updates were applied to the non-repeated part and a portion of the main body layers, with FedTLU selecting layers based on scores and Random selecting layers uniformly at random. The Last update case involved updating only the last few layers of the non-repeated part. Importantly, for FedTLU, Random, and Last updates, the same number of layers were updated in each experiment to ensure a fair comparison.

Each case was evaluated three times, and the average perplexity, which measures prediction accuracy (\textit{lower is better}), was reported. Global testing perplexity was calculated after each aggregation using test sets from the datasets, while local testing performance was measured by averaging perplexity scores from participating clients after the local training using the same test sets.

\subsection{Experimental Results}
We compared FedTLU's performance under three conditions: \textbf{1)} training the global model from scratch, \textbf{2)} fine-tuning pre-trained models, and \textbf{3)} scenarios with noisy or potentially malicious clients.

\textbf{For training from scratch}, each communication round varied the portion of updated layers (75\%, 50\%, and 25\%). Each local client was allocated up to a maximum number of tokens, calculated by dividing the total dataset size by the number of clients, where tokens refer to the units obtained from tokenizing the dataset. 
A minimum number of tokens is half of this maximum number. Local datasets were non-overlapping, and 10\% of clients participated per round. Table I shows that FedTLU consistently achieved better test perplexity, reducing global perplexity by up to 1.14\% and local perplexity by up to 1.3\% compared to Random. FedTLU also showed a 41.7-120.8\% improvement in global perplexity and 13.8-116.2\% in local perplexity compared to last-layer updates. These results indicate that targeted layer updates in FedTLU enhance model stability.

\textbf{For fine-tuning}, we trained the global model for 400 rounds with full updates. The same token distribution approach as in training from scratch was applied, ensuring consistency across experiments.
During fine-tuning, 5\% of clients participated per round, and each client’s local data was halved. The update portion was decreased sequentially from 75\% to 50\% to 25\%, and if the loss did not decrease by at least 1\% for 10 consecutive rounds, the portion was further reduced. Table II and Fig. 1 show that FedTLU outperformed other strategies, reducing global perplexity by 2.54-3.50\% compared to full updates, by up to 0.97\% compared to Random, and by 3.18-3.80\% compared to last-layer updates. Locally, FedTLU achieved reductions of 2.60-3.15\%, 0.06-0.92\%, and 3.23-3.36\%. In Fig. 1, FedTLU consistently maintained lower global perplexity during training.

\textbf{In noisy or malicious client scenarios}, 10\% of clients were assigned shuffled labels (i.e., mismatched tokens). Table III shows that FedTLU reduced global perplexity by 2.13-2.99\% compared to full updates and by -0.23-7.86\% compared to Random. Locally, FedTLU achieved reductions of 1.88-2.67\% and -0.17-8.27\%. \textit{Notably, Random increased perplexities in the GPT-2 (UDPOS) case from the full update cases, highlighting that randomly selecting layers may include less effective ones}. These results emphasize the importance of selective layer updates. In summary, FedTLU demonstrated better resilience, reducing the impact of noisy or malicious clients and maintaining lower perplexity, providing robustness in federated learning environments. As seen in Fig. 2, FedTLU consistently maintained lower global perplexity despite noisy updates.
\section{Conclusion}\label{sec:conclusion}
We introduced FedTLU, a targeted layer update strategy for federated learning (FL). FedTLU improves convergence and reduces the impact of noisy client updates. Experimental results show that FedTLU outperforms random and last-layer updates, providing a more efficient and stable solution for FL in language modeling.

\bibliographystyle{IEEEtran}
\bibliography{mybib}
\end{document}